\documentclass[11pt]{article}

\usepackage[utf8]{inputenc}
\usepackage[T1]{fontenc}
\usepackage[margin=1in]{geometry}
\usepackage{hyperref}
\usepackage{url}
\usepackage{booktabs}
\usepackage{amsfonts}
\usepackage{amsmath}
\usepackage{amssymb}
\usepackage{nicefrac}
\usepackage{microtype}
\usepackage{xcolor}
\usepackage{natbib}
\usepackage{authblk}

\newif\ifrev
\revfalse
\ifrev
  \newcommand{\rev}[1]{\textcolor{magenta}{#1}}
  \newcommand{\cfix}[1]{\textcolor{green!50!black}{#1}}
  \newcommand{\tidy}[1]{\textcolor{blue!60!black}{#1}}
  \newcommand{\drift}[1]{\textcolor{cyan!85!black}{#1}}
  \newcommand{\bridge}[1]{\textcolor[HTML]{ED7D31}{#1}}
  \newcommand{\viola}[1]{\textcolor[HTML]{7030A0}{#1}}
\else
  \newcommand{\rev}[1]{#1}
  \newcommand{\cfix}[1]{#1}
  \newcommand{\tidy}[1]{#1}
  \newcommand{\drift}[1]{#1}
  \newcommand{\bridge}[1]{#1}
  \newcommand{\viola}[1]{#1}
\fi

\title{AI Evaluation Should Measure Verification Cost, Not Correctness Alone\thanks{Preprint, August 2026. \copyright{} 2026 The Authors. Licensed under the Creative Commons Attribution 4.0 International License (CC BY 4.0)}}

\author[1]{Viviana Crescitelli}
\author[2]{Generoso Immediato\thanks{Corresponding author: \texttt{generoso.immediato@hitachirail.com} --- ORCID: \href{https://orcid.org/0009-0006-0294-6264}{0009-0006-0294-6264}.}}
\author[3]{Fabio Persia}
\author[3]{Stefania Costantini}
\affil[1]{Hitachi, Ltd., Tokyo, Japan --- \texttt{viviana.crescitelli.mh@hitachi.com}}
\affil[2]{Hitachi Rail, Naples, Italy}
\affil[3]{Department of Information Engineering, Computer Science and Mathematics, University of L'Aquila, Italy --- \texttt{\{fabio.persia, stefania.costantini\}@univaq.it}}

\date{August 2026}

\begin{document}

\maketitle

\begin{abstract}
The reliability of AI generative models is typically measured by output correctness, yet in practice it depends on the effort required to verify those outputs. We argue that current evaluation metrics overlook a critical failure mode: Verification-Cost Errors (VCEs), \viola{defined as incorrect input-output pairs that a declared fraction of the verifier population fails to identify within the verification budget available in a given deployment context}. \viola{Unlike standard notions of ``hallucination'', VCEs are defined operationally, by the failure of correct identification within budget rather than by any property of the output itself. Plausibility and authoritative presentation are hypothesised contributors to that failure, not defining conditions.} To capture this asymmetry, we introduce the notion of verification cost relative to a deployment budget \viola{as an operational dimension that current evaluation does not routinely capture}. The quantity is presented as a conceptual instrument rather than a finalized metric. Evidence from code generation and multi-modal document understanding shows that high benchmark accuracy can mask significant verification effort in practice. We therefore take the position that correctness alone is insufficient as a measure of reliability. AI evaluation should explicitly account for verification cost, reflecting whether errors can be detected under realistic resource constraints.
\end{abstract}

\section{Introduction}

Current evaluation of AI generative models is misaligned with real-world deployment risk. In fact, standard benchmarks focus on accuracy, but reliability in practice depends on the effort required to verify the model's results.
Consequently, models that perform well in benchmark tests may still, in practice, entail a significant verification burden.
For instance, consider a code-generation model that produces a sorting function. The function passes all the provided tests, but fails in edge cases such as empty inputs or duplicate values that are not covered by the test suite. Generating the code takes seconds, but detecting the error requires constructing additional tests or performing manual inspection. In this setting, the cost of verification exceeds the cost of generation; however, unfortunately, this asymmetry is not captured by standard evaluation metrics.
\cfix{Furthermore, several strands of prior work identify failure modes that can plausibly generate substantial downstream validation demands --- AI safety failures, foundation-model risks, factual falsehoods, hallucinations, and failures of multi-step reasoning \citep{amodei2016concrete, bommasani2021foundation, lin2022truthfulqa, li2023halueval, creswell2023selection} --- although these works do not generally measure verification cost directly;}
as a result, we argue that the central limitation of current evaluation is not error frequency, but verification cost.

Existing work has identified limitations in current evaluation practices, including distributional shifts, annotation artifacts, and data contamination \citep{liang2023helm, kiela2021dynabench, deng2024contamination}. \cfix{Research on hallucinations further shows that models can produce plausible continuations that are not factually correct \citep{lin2022truthfulqa, li2023halueval}.} \cfix{\viola{Complementary work indicates} that explanations provided by AI can lead to inappropriate reliance and do not necessarily improve the combined performance of humans and AI \citep{bansal2021does},} and that generative systems shift the burden of verification onto users operating under constraints of time, expertise, and attention \citep{sariyar2026gecos, Immediato2025-POM}.
\tidy{Taken together, these observations point to a dimension that current evaluation omits: verification cost. Protocols that ignore it are incomplete, because reliability depends not only on correctness but on whether errors can be detected under realistic resource constraints.}
\rev{\textbf{Our position is that AI evaluation should explicitly measure verification cost relative to a deployment verification budget, rather than correctness alone.}}
More specifically, this paper points out what follows:
\begin{itemize}
    \item reliability cannot be measured by correctness alone; it must also take verification costs into account;

    \item a critical class of failures arises when errors are difficult to detect under realistic constraints;

    \item evaluation should explicitly measure the effort required to verify the model outputs.

\end{itemize}
To formalize this perspective, we introduce \viola{a single construct}:
\textbf{Verification-Cost Errors (VCEs)}, \viola{defined as incorrect input-output pairs that a declared fraction of the verifier population fails to identify within the verification budget available in a given deployment context}.
The remainder of this paper is organized as follows. Sections~\ref{sec:broken}--\ref{sec:limits} diagnose the gap in current evaluation; Section~\ref{sec:cognitive-problem} frames verification as a computational and cognitive problem; Section~\ref{sec:concept} \viola{formalizes VCEs together with the success-probability and observed-burden measures}; Section~\ref{sec:method} proposes a verification-aware benchmarking methodology; Sections~\ref{sec:implications}--\ref{sec:concludingremarks} discuss implications, alternative views, and limitations.

\section{What is Broken: Verification-Blind Evaluation}
\label{sec:broken}

Current evaluation protocols often operationalize correctness as agreement with reference answers, while leaving the cost of establishing that correctness largely unmeasured \citep{liang2023helm,kiela2021dynabench,gururangan2018annotation,deng2024contamination}.

A critical class of failures occurs when outputs are \emph{plausible}, \emph{consistent with the context}, and \emph{difficult to refute}.
At first glance, these results appear to be correct, but verifying them requires considerable effort.
This creates a mismatch between benchmark assumptions and real-world usage, where users must verify outputs under limited time and expertise constraints. As a result, evaluation pipelines systematically underestimate the risk of failure.

\viola{Therefore, we focus throughout the paper on a subclass of interest, characterised along three observable dimensions: outputs that are (i) \emph{stylistically fluent}, (ii) \emph{locally coherent}, and (iii) \emph{authority-mimicking} --- that is, they reproduce the rhetorical and formatting conventions of expert discourse without satisfying its constraints. These properties are hypothesised to raise the probability that verification fails within budget; they do not enter the definition given in Section~\ref{sec:concept}, which is satisfied equally by an error that escapes detection because the task is intrinsically hard, because the available verifiers lack the relevant expertise, or because the tooling is inadequate.} \viola{We next give the broader VCE class its operational definition in Section~\ref{sec:concept}:}
\viola{a \emph{Verification-Cost Error} is any incorrect input-output pair that a declared fraction of the verifier population fails to identify within the verification budget available in a given deployment context.} Unlike qualitative typologies of model failure, \rev{this cost-asymmetric framing admits measurable proxies within the protocol proposed in Section~\ref{sec:method}.}

\textbf{The deceptive nature of fixed benchmarks is most apparent in code generation}. \tidy{Models can exploit the sparse coverage of standard test suites:} \cfix{since benchmark performance within evaluated distributions does not guarantee robust generalization to omitted or out-of-distribution cases~\citep{liang2023helm}, generated code may appear ``apparently correct'' while retaining latent failures} \tidy{in the boundary cases that automated evaluation omits.}
Identifying these bugs requires the user to construct exhaustive test cases or perform a manual, line-by-line inspection --- a verification effort that quickly exceeds the generation time, which is measured in fractions of a second.
In production environments where exhaustive testing is infeasible, benchmark accuracy serves as a misleading indicator of reliability.

\textbf{Multi-modal documents} understanding exhibits a parallel pathology: the ``smoothing'' of sensor noise into plausible but semantically distorted text. In OCR--LLM pipelines, minor character-level errors can be corrected by the language model, transforming sentences into syntactically valid but factually incorrect clauses---for instance, converting a conditional legal obligation into an unconditional one.
Since the resulting output remains contextually consistent, identifying the error requires a careful manual comparison with the original source.
Current extraction metrics, by focusing on character or word-level accuracy, fail to capture this semantic volatility and the resulting tax on human oversight.

\textbf{Retrieval-Augmented Generation (RAG) }often exacerbates this asymmetry by providing a false veneer of grounded authority. While grounding is intended to improve reliability, models frequently misrepresent the retrieved content, by reversing causal relationships or overgeneralizing the results.
\cfix{RAG outputs may thus require users to verify not only the generated claim but also whether the cited or retrieved evidence actually supports it --- an additional claim--source alignment task. Related evaluation and tool-augmentation work illustrates how factual claims may be decomposed into atomic units or supported through external computation \citep{min2023factscore, gao2023pal}, but these mechanisms, in and of themselves, do not eliminate the need for downstream validation.}
These failures are compounded by human cognitive constraints; \cfix{in fact, evaluators operating under limited attention may over-rely on outputs presented in a well-founded and authoritative style~\citep{bansal2021does}.}
Consequently, grounding mechanisms do not eliminate the verification gap;
\rev{they can instead hide it behind a facade of credibility, as the evidence below illustrates.}
To this extent, \cite{magesh2025hallucinationfree} conducted \cfix{the first preregistered evaluation} of proprietary RAG-based legal AI tools and found that systems marketed as ``hallucination-free'' continue to generate hallucinations in 17\% to 33\% of cases.
Essentially, the presence of retrieval grounding did not eliminate the verification burden,
but rather restructured it, requiring users to assess the accuracy of citations rather than the truthfulness of statements themselves.
Similarly, \cite{dahl2024legalfictions} profiled legal hallucinations across four major LLMs and documented a systematic tendency to counterfactual bias and overly confident responses --- precisely the conditions that maximize verification cost.
\viola{Direct measurement of the resulting burden is rarer, but not absent. In a randomized controlled trial, \citet{becker2025metr} found that experienced open-source developers took 19\% longer to complete real tasks when AI assistance was permitted, while estimating afterwards that it had made them 20\% faster. The finding is specific to the tooling of early 2025 and the authors present it as such; what generalises is the shape of the result, namely a measured cost that runs opposite to the perceived one --- which is the condition our framework is designed to make visible.}
\rev{We note that these figures are hallucination rates rather than VCE rates: they document the practical setting that motivates our framework, not estimates of the quantities defined in Section~\ref{sec:concept}.}

{To sum up, the patterns highlighted}
{in Table~\ref{tab:vbi_asymmetry} suggest that verification-cost errors \viola{may not be isolated anomalies, but recurrent features of current generative and retrieval-augmented architectures}. \cfix{While scaling and capability studies document substantial gains in performance and apparent competence \citep{wei2022emergent, bubeck2023sparks}, they provide no direct evidence of a commensurate reduction in the effort required to falsify model outputs.}
In fact, benchmark gains may not result in a proportional reduction in the real-world labor required for certification.}
\begin{table}[t]
\centering
\small
\begin{tabular}{lccc}
\toprule
\textbf{Task} & $C_g$& $C_v$& \textbf{Typical Failure Mode}\\
\midrule
Code Generation & Low & High & Hidden edge-case bugs\\
Factual QA & Low & High & Plausible but false claims\\
RAG Systems & Medium & High & Misleading grounded outputs\\
Document Understanding & Low & High & OCR-induced semantic distortions\\
\rev{Schema-constrained extraction} & \rev{Low} & \rev{Low} & \rev{Violations surface on automated validation}\\
\rev{Certificate-producing generation} & \rev{Low} & \rev{Low} & \rev{Outputs carry machine-checkable certificates}\\
\bottomrule
\end{tabular}
\caption{Illustrative asymmetry between \textit{generation cost} ($C_g$) and \textit{verification cost} ($C_v$). \viola{The regime $C_v \gg C_g$ is what motivates the framework; the formal condition defining a Verification-Cost Error is given in Section~\ref{sec:concept} and is stated in terms of verification outcomes, not of this comparison.} \rev{The last two rows illustrate the opposite scenario, in which an efficient decision-making process makes verification cheap (possibly $C_v < C_g$): the asymmetry is a risk class, not a universal law.}}
\label{tab:vbi_asymmetry}
\end{table}

\section{Limits of Current Framing}
\label{sec:limits}

{Current reliability frameworks---hallucination, calibration, abstention, selective prediction, and benchmark robustness---only partially capture verification cost and rarely treat it as a first-class metric. They typically assume that errors are obvious upon inspection, an assumption that fails for a critical class of outputs: still, those that are \emph{plausible}, \emph{internally consistent}, and \emph{deceptively difficult to falsify}.}
\tidy{We consider five such framings in turn --- hallucination, calibration, benchmarking, explanation, and tool augmentation --- and show that each of them improves a certain aspect of reliability, without, however, quantifying the cost required to reach a conclusion. The recurrence of that gap across otherwise unrelated framings is itself the argument: it is pervasive rather than accidental.}
\rev{Verification cost is best read as one measurable axis within the broader landscape of AI reliability \citep{rabanser2026reliability}: holistic evaluation efforts for models \citep{liang2023helm} and agents \citep{kapoor2026hal} already move beyond binary success metrics, yet none treats the cost of establishing a verdict as a first-class quantity. Our claim is accordingly narrower than a theory of reliability: verification cost is a missing, measurable dimension of it.}
{We retain the term \emph{hallucination} for continuity with the established literature, but use it cautiously: the term captures the surface phenomenology of a plausible but defective output, while leaving under-specified the operational question central to this paper, namely whether the failure can be detected within realistic verification constraints. This concern is consistent with the JORABP taxonomy proposed by \citet{Immediato2025-POM}, which distinguishes \emph{Juicy Oranges}---valid and reliably verifiable outputs; \emph{Rotten Apples}---clearly incorrect outputs whose invalidity is readily evident; and \emph{Banana Peels}---incorrect but highly plausible outputs whose detection imposes substantial epistemic and practical verification burdens. The present paper later formalizes the Banana Peel concept through the VCE condition introduced in Section~\ref{sec:concept}:
\viola{an incorrect model response becomes deployment-relevant when a declared fraction of the verifier population fails to identify it within the available verification budget.} \viola{Within the plausibility-focused subset described in Section~\ref{sec:broken}, and not for the VCE class as a whole, the classification summary may serve as an operational proxy for the residual Banana Peel prevalence observed downstream of the generative system.}}

\textbf{Hallucination vs. Detectability.} Defining failures as ``hallucinations'' is insufficient, since it prioritizes \textit{incorrectness} over \textit{detectability}.
\cfix{Although the literature focuses primarily on detecting factual errors \citep{lin2022truthfulqa, li2023halueval, manakul2023selfcheckgpt}, incorrect outputs may remain superficially plausible and difficult to distinguish from correct ones; the bottleneck thus shifts from error identification to external validation.}
Hallucination frameworks tell us whether a model is incorrect, but they say nothing about {the effort required of the user to prove it.}
The metaphor itself deserves thorough analysis.  \cite{forster2025factfairy} \cfix{trace} the genealogy of ``\textit{hallucination}'' from its origins in computer vision—where it denoted the deliberate process of enhancing blurry photographic images—to its current deployment as an under-specified catch-all in NLP discourse.
They argue that the metaphor anthropomorphizes models as ``mad or infantile subjects'', simultaneously normalizing non-factual output and deflecting responsibility from designers onto the system itself.
This discursive function compounds the verification problem: by framing errors as quasi-psychological phenomena, the hallucination metaphor obscures the structural cost asymmetry; {to cope with this issue, in this paper we define the \emph{Verification-Cost Errors} (\emph{VCEs}, Section~\ref{sec:concept}). }
{In our perimeter, what matters operationally is not whether an output is a ``hallucination'' in some cognitive sense, but whether it can be falsified within the user's resource budget.}

\textbf{The Limits of Calibration.} {Calibration and uncertainty estimation offer only partial relief. While these methods align confidence with accuracy \citep{guo2017calibration, jiang2021calibration} and can help prioritize where verification effort is allocated, they rely on an epistemic self-assessment that models often lack, and they do not lower the verification threshold itself.} High-stakes outputs still demand exhaustive audit {regardless of the level of reliability indicated by the model;}
calibration merely {alerts us} when we should be suspicious, without reducing the labor required for certainty. Probabilistic methods for detecting unreliable outputs represent a partial advance.
{In such a context,} \cite{farquhar2024semanticentropy} \cfix{introduce} \textit{semantic entropy}---an entropy computed over meanings rather than token sequences---to detect a specific subclass of hallucinations they term \textit{confabulations}: outputs that are both wrong and arbitrary. While effective within its scope,
{the method does not explicitly account for cases in which large language models (LLMs) make errors systematically and with confidence.}
This distinction is significant: it confirms that detectability depends on the \textit{type} of failure mechanism, not merely on the presence of error---a principle that VCEs generalize by centering the cost of detection as the primary metric.

\textbf{Benchmark blindness.} Benchmarks {hide} verification asymmetry by treating all errors as uniform. Standard evaluations measure agreement with references, \citep{liang2023helm, kiela2021dynabench} {but generally do not record the labor required to reach that agreement}.
Even granular methods, such as atomic fact verification \citep{min2023factscore}, paradoxically expand the {scope of} verification by forcing users to {evaluate} a {fragmented} list of sub-claims.
{Benchmark gains may therefore not translate proportionally into reductions in human oversight cost.}

\textbf{The Explanatory Tax.} We argue that explicit reasoning and explanations may actually worsen the verification burden. \cfix{Chain-of-thought explanations need not faithfully reflect the process that produced the final answer \citep{dziri2023faithful},}
{yet they provide a semblance of logic that leads users to rely on them too heavily}~\citep{bansal2021does}.
{As models scale \citep{kaplan2020scaling, wei2022emergent, bubeck2023sparks}, the perceptual gap between correct and deceptive outputs may narrow.} In this {context}, \viola{interpretability may function as reassurance rather than as a diagnostic instrument.}
{Verification burden is also shaped by human cognitive constraints, including time pressure, attention limits, and \cfix{explanation-induced overreliance} \citep{bansal2021does}, which affect the verification capacity of human overseers in ways that current evaluation does not capture.}

\textbf{Tool-Augmented Shifting.} \viola{Finally, tool augmentation may relocate the verification cost rather than remove it.}
\cfix{Although external tools can improve accuracy in structured tasks \citep{gao2023pal}, we argue that they may also introduce additional artifacts --- programs, executions, retrieved evidence, or tool outputs --- that must themselves be validated.}
This creates a multi-stage verification recursive loop. \cfix{Canonical treatments of deep learning and causal representation learning focus primarily on learning, generalization, and causal structure \citep{goodfellow2016deep, scholkopf2021causal}, rather than on the downstream human cost of validating individual outputs.}
\cite{Immediato2025-POM} argues that the co-pilot metaphor may install inflated expectations of AI competence while obscuring the verification responsibility that remains with the user. Unlike an aviation co-pilot—trained to the same standard, accountable under the same regulatory regime—an AI ``co-pilot'' is probabilistic, non-transparent, and incapable of assuming liability.
The metaphor thus performs precisely the kind of trust inflation that maximizes the conditions for Verification-Cost Errors: users lower their verification effort in response to a framing that suggests shared competence.

\section{Verification as a Computational and Cognitive Problem}
\label{sec:cognitive-problem}

Verification cost is a structural byproduct of the mismatch between generative sampling and formal validation.
While model outputs are produced via low-cost forward inference, verification necessitates fundamentally different operations: backward reasoning, combinatorial search, or external grounding.
This creates an intrinsic asymmetry
{in which generating a response is a computationally trivial task, whereas verifying it is not.}

{Let us consider the \text{Input} and \text{Output} sets.
Then, $\forall x \in \text{Input}$, and
$\forall y \in \text{Output}$:}
\begin{itemize}
    \item {$C_g(x,y)$ denotes the \emph{generation cost};}
    \item {$C_v(x,y)$ denotes the \emph{cost of determining correctness}.}

\end{itemize}
{In deployment settings, reliability is governed by the regime:}
\begin{equation}
    C_v(x,y) \gg C_g(x,y).
\end{equation}
Where $\gg $ denotes a \textbf{regime distinction} rather than a fixed numerical threshold, indicating that verification costs dominate generation costs under realistic resource constraints.

{Thus, such a} \textbf{Verification Gap} implies that an agent's utility is limited not by its generative capacity, but rather by what the user can afford to audit.
Natural language scales this imbalance: generating plausible prose is a low-latency sampling process, whereas verification demands multi-level retrieval or domain-specific expertise.

We refer to the activity just described as \textit{Cognitive Verification}: a form of epistemic validation in which humans do not certify the system itself, but rather evaluate the meaning, trustworthiness, and applicability of a specific machine-generated output \citep{Immediato2025-POM}.
\cfix{Cognitive verification often recruits the deliberative processes associated with System 2} \citep{kahneman2011} and echoes the symbol-grounding problem \citep{harnad1990}: models may lack the grounding that the user is called to supply, {thereby shifting the burden of contextual validation and referential stabilization onto the individual.}

\rev{This burden falls on human verifiers whose time, attention, and expertise are limited (Sections~\ref{sec:broken}--\ref{sec:limits}); analyses based on systems theory sharpen the point.}
\cite{sariyar2026gecos} proposes that LLMs function as \textit{operators for general cognitive shortcuts}---systems that stabilize communicative continuation (i.e., prioritizing the likelihood of the next token over semantic truth) by producing contextually plausible outputs without epistemic commitment. Responsibility does not disappear in such an interaction; it \textit{migrates} from the generator to the user and to the \textbf{socio-technical infrastructure surrounding the output.} This redistribution is structurally obscured by conversational interfaces that invite anthropomorphic interpretations. When \textit{connectability} substitutes for justification, the result is what \cite{sariyar2026gecos} {refers to as} \textit{normative flattening}: fluency and stylistic adequacy are privileged over epistemic depth, thereby creating precisely the conditions under which \textbf{Verification-Cost Errors} proliferate.

{The relationship between model capability and verification cost is empirically unsettled. Greater fluency may make errors harder to detect by reducing surface markers of incorrectness; conversely, more capable models may produce outputs whose internal consistency aids verification (e.g., by yielding cleaner intermediate representations). We conjecture, but do not establish, that current scaling trends widen rather than narrow the verification gap \citep{wei2022emergent, bubeck2023sparks}, and that this is in itself an issue that verification-aware benchmarking could resolve. What is uncontroversial is the structural point: reliability is bounded not only by correctness, but also by the feasibility of verification under resource constraints,} \viola{and this limitation is not routinely captured by current evaluation methods.}

\section{Proposed Concept: Verification Cost}
\label{sec:concept}
Building on this asymmetry, we argue that evaluation must treat verification as a first-class metric rather than a binary assumption. {Our novelty claim is narrow and deliberate. We identify a cross-cutting evaluation category: known failure modes---hallucinations, miscalibrated confidence, unfaithful chain-of-thought, and others---\viola{become operationally significant when a declared fraction of the verifier population cannot correctly identify them within the available verification budget}. We do not claim a new failure phenomenon. \viola{Nor do we claim that evaluation has ignored cost as such: enterprise-oriented frameworks already report execution cost, latency and reliability alongside accuracy \citep{mehta2025clear}, and programmes such as NIST GenAI study the gap between generation and automated detection \citep{nist2024genai}. Those quantities are machine-side. \viola{What we operationalize is deployment-relative verification \emph{success} and observed \emph{human} verification effort, reported as separate evaluation dimensions and treated as measurable indicators of an otherwise latent verification cost. The distinction matters throughout: the cost of establishing correctness is the target of interest, while what the protocol observes is whether correctness was established within budget and how much effort the attempt absorbed.}} Verification cost is orthogonal to existing failure taxonomies in the sense that any of them can have low or high verification cost, and it is the cost dimension that determines whether known failures translate into deployment risk.}

\textbf{{Preliminary Definitions}.}
{Given the following sets and possible items:}
\begin{itemize}
    \item {\emph{CU} as the set of \emph{Cost Units} in which both \emph{generation} and \emph{verification} efforts can be measured;}
    \item {depending on the considered domain, possible items $u \in CU$ are \emph{human-minutes}, \emph{FLOPs}, or \emph{USD}};
    \item {$\mathcal{V}$ as the set of task- and deployment-dependent \emph{Verifiers}, equipped with a probability distribution that reflects the population of verifiers operating in the deployment context;}
    \item {possible elements $v \in \mathcal{V}$ are \emph{human experts}, \emph{audit procedures}, or \emph{automated validation protocols}. Throughout the paper, expressions of the form $\mathbb{E}_{v \sim \mathcal{V}}[\cdot]$ denote expectations under this distribution.}
    \item \viola{In the protocol developed in Section~\ref{sec:method}, however, $\mathcal{V}$ denotes a population of \emph{human} verifiers. Several of its requirements --- self-reported confidence as a stopping criterion, the option to declare an output unverifiable, verifier-experience metadata, agreement across verifiers --- are specific to human-subject measurement, and the dimension the paper claims to operationalize is human verification effort. Automated validation procedures can be studied under an adapted protocol, but that adaptation lies outside the present operationalization.}

\end{itemize}
{We provide the following preliminary definitions.
$\forall x \in \text{Input}$, and
$\forall y \in \text{Output}$:}
\begin{itemize}
    \item {\mbox{$C_g^{u}(x,y)$} denotes the cost of generation in unit $u \in CU$;}
    \item {\mbox{$C_v^{u}(x,y;v)$} denotes the cost of verifying $y$ under verifier \rev{$v \in \mathcal{V}$}}.
    \item \viola{Two distinct quantities are latent in that phrase, and we separate them. The \emph{decision time} is the effort expended until the verifier commits to a verdict, whatever that verdict turns out to be; the \emph{correct-verdict time} is the effort expended until a verdict agreeing with the ground truth is reached. We take \mbox{$C_v^{u}$} to denote the second: it is the cost of \emph{determining correctness}, which is what the framework is about. The two coincide whenever the verdict is correct, and diverge exactly on the episodes discussed in Step~3 of Section~\ref{sec:method}. Where they diverge, \mbox{$C_v^{u}$} is not observed and is not recoverable: a verifier who stops early with a mistaken verdict has told us nothing about how long a correct one would have taken. We therefore report \mbox{$C_v^{u}$} only over the episodes in which a correct verdict was in fact reached, and carry the remaining episodes as outcomes in their own right rather than as imputed costs.}

\end{itemize}
\rev{For human-in-the-loop deployments, verification cost may be measured in $u=\text{human-minutes}$ over a specified population of human verifiers, since deployment reliability is ultimately bounded by human oversight capacity (Section~\ref{sec:cognitive-problem}); for fully automated pipelines, $u=\emph{FLOPs}$ or $u=\emph{wall-clock seconds}$ may instead be appropriate. All formal constructs in this section are defined relative to a chosen $u$.}
\rev{More generally, the verification cost is not an intrinsic property of a model: it is a property of a model \emph{output} evaluated according to a specific measurement specification $(u,\mathcal{V},B,\tau,\viola{q})$, with interface, tools, and protocol held fixed. All comparative claims in this paper are relative to such a specification.}
\rev{In addition, rather than assuming a unique target output, we define an \emph{acceptance relation} $\mathcal{A} \subseteq \text{Input} \times \text{Output}$, where $(x,y)\in\mathcal{A}$ means that $y$ is an acceptable output for input $x$; tasks admitting multiple correct outputs (e.g., program synthesis, where the same specification is satisfied by many programs) are thereby covered. Given a model output function \viola{$g$, understood as a realization of a generative system under a fixed configuration --- prompt, decoding parameters and system snapshot --- rather than as a deterministic map, since the same input may yield different outputs across runs}, we define the set of incorrect input-output pairs as follows:}
\rev{
\[
\mathrm{IncorrectPairs}
=
\{(x,\hat{y}) \in \text{Input} \times \text{Output} \mid \hat{y}=g(x),\ (x,\hat{y})\notin\mathcal{A}\}.
\]
}
\rev{Correctness is thus a property of the input--output \emph{pair}, not of the output in isolation: the same output $\hat{y}$ may be acceptable for one input and unacceptable for another.}

\textbf{Verification-Cost Errors (VCEs).} {Verification cost is meaningful only in relation to a deployment verification budget. Let $B$ denote the verification budget available in a given deployment context, expressed in the same unit $u$ (e.g., the number of human-minutes a clinician, lawyer, or developer can realistically allocate per output).
\viola{An earlier formulation declared a pair a Verification-Cost Error when its expected verification cost exceeded $B$. We do not retain that form, because a protocol that halts at the budget never observes by how much the cost exceeds it, so the condition is not decidable from the evidence the protocol collects. We define the class instead on the outcome the protocol does observe. Let $S_B(x,\hat{y};v)$ be the indicator that verifier $v$ reaches a verdict agreeing with the ground truth within the budget. Then, for a pre-registered threshold $q$,}
\[
(x,\hat{y}) \in \mathrm{VCE}_{B,q}
\iff
(x,\hat{y}) \notin \mathcal{A}
\;\wedge\;
\Pr_{v\sim\mathcal{V}}\!\left[S_B(x,\hat{y};v)=0\right] \ge q .
\]
\viola{In words: the pair is incorrect, and at least a fraction $q$ of the verifier population fails to establish that within the available budget. The definition is stated over the population, whereas an instantiation observes a finite sample of verifiers, so the empirical rule must be declared with the rest of the specification: the decision rule is pre-registered along with the estimator, and the full measurement specification accordingly comprises \mbox{$(u,\mathcal{V},B,\tau,q,m,\text{interval method},\alpha,\text{decision rule},\text{tools},\text{interface},\text{dataset})$}. \viola{Where we write $P$ below we mean this specification in full; the shorter tuples are abbreviations for it.} With the three verifiers of Step~4 the attainable proportions are coarse --- $0$, $\tfrac13$, $\tfrac23$, $1$ --- and a rule resting on the point estimate alone would be brittle. We therefore recommend a three-state classification: a pair is a Verification-Cost Error when the lower confidence bound on \mbox{$\Pr[S_B=0]$} reaches $q$, it is not one when the upper bound falls below $q$, and it is recorded as \emph{inconclusive} otherwise. Whichever rule is adopted, the estimate, its uncertainty interval and the number of verifier observations are part of the reported result rather than an appendix to it, and the interval method and confidence level are declared with the rest of the specification.
Two consequences must be faced rather than glossed. The first is that the three verifiers required in Step~4 are not enough to classify individual outputs: with $q=\tfrac12$ and conventional two-sided $95\%$ binomial intervals, three failures out of three yield a lower bound near $0.29$ and no failures out of three an upper bound near $0.71$, so neither extreme resolves and every output is recorded as inconclusive. The requirement $N\ge 3$ supports agreement reporting; the number of verifiers needed for output-level classification is a separate quantity, to be derived from the interval method, the confidence level, the threshold $q$ and the proportion of inconclusive results one is willing to tolerate. \viola{Under the same conventions, resolution at $q=\tfrac12$ begins at six verifiers per output for the two extreme outcomes --- six failures out of six give a lower bound of about $0.54$, and none out of six an upper bound of about $0.46$ --- although substantially larger samples are needed to control the overall share of inconclusive results when the underlying probability lies near $q$.}
The second is that a three-state classification does not yield a single rate. We therefore report the confirmed rate and the inconclusive rate over the same denominator, the incorrect outputs, and report the resulting \emph{classification envelope}:}
\[
r_{\mathrm{lower}} = r_{\mathrm{confirmed}}, \qquad
r_{\mathrm{upper}} = r_{\mathrm{confirmed}} + r_{\mathrm{inconclusive}} .
\]
\viola{This classification envelope summarises the output of the pre-registered decision rule: it is descriptive, not inferential. Because the per-output intervals carry no simultaneous coverage guarantee, a confirmed classification may be a false positive and a rejected one a false negative, so the envelope is neither a confidence interval for the true rate nor a bound that provably contains it; a guarantee about the underlying rate would require simultaneous inference or a hierarchical model over outputs, which we do not attempt here. Reporting the pair rather than a point value is nonetheless the honest form of the result when verifier coverage is thin, since it makes that thinness visible instead of hiding it in a rounded proportion.} The threshold is a deployment parameter like $B$ itself and is declared, not universal; $q=\tfrac{1}{2}$ reads as ``at least half of the verifier population would not catch it''. Nothing in this definition requires knowing how long verification would have taken had it continued.}
{This formulation has three advantages. First, it eliminates the undefined relation $\gg$ in favor of a bounded comparison.
Second, it makes VCE deployment-relative: \rev{the same input-output pair may cross the VCE threshold as the available verification budget changes, holding the task, verifier population, interface, and cost unit fixed --- e.g., a 5-minute versus a 60-minute review budget within the same clinical deployment ---} which matches operational reality.
\viola{Third, it yields observable success and burden measures, which are the primary indicators of deployment risk in this paper.}}
\rev{This licenses a two-level reading. The occurrence of an \emph{individual} VCE is a relational property of the pair $(x,\hat{y})$ under the declared measurement specification; it is not a property of the model. \viola{Within the plausibility-focused subclass of Section~\ref{sec:broken}, such an event corresponds to a Banana Peel in the taxonomy of Section~\ref{sec:limits}; outside it, a VCE may arise from task difficulty or from the limits of the verifier population without having any such character.} \viola{The VCE \emph{classification summary} --- confirmed rate, inconclusive rate and envelope --- measured under a fixed protocol} $P=(u,\mathcal{V},B,\tau,\viola{q},\text{tools},\text{interface},\text{dataset})$ is, by contrast, a comparative property of the model--protocol pair, in exactly the sense in which standard benchmarks attribute scores to models: under protocol $P$, \bridge{a given system} exhibits \viola{a confirmed VCE rate, an inconclusive rate, and the corresponding classification envelope}. When dataset, verifier population, budget, tools, and protocol are held identical, differences in observed rates across models \viola{are conditionally associated with the evaluated system under that fixed protocol --- residual variability from sampling, from run-to-run variation and from output--verifier interaction remains --- and are therefore reported as model-level comparative indicators} (Section~\ref{sec:method}, Step 6); no external validity across different protocols is implied.}

\drift{\textbf{Temporal validity.} The relativisation just described is synchronic, and one further consequence bears on how the quantity should be used. Several constituents of a measurement may drift over time: verifier populations accumulate experience with a system's characteristic outputs, verification tooling improves, and budgets shift under organisational pressure. On the generation side, a deployed system is a service rather than a fixed artefact, since weights, decoding parameters, and system-level prompts may change while its name does not. A declared specification is therefore a specification \emph{at a time}, and a verification-cost measurement carries a shelf life: direct comparison between systems requires either a common measurement window or a design that explicitly controls for changes in the evaluated system and in the protocol. It also follows that the system under evaluation must be kept distinct from the apparatus that evaluates it --- the generation-side configuration is recorded alongside the protocol, not absorbed into it --- since otherwise a comparison between two systems would become a comparison between two protocols.}

\drift{A related methodological risk deserves note. Evaluation artefacts may re-enter training corpora \citep{deng2024contamination}, and preference-based optimisation adapts generation to recurring patterns of human judgement. Verification cost may therefore exhibit co-adaptive dynamics, with both the evaluated system and the verifier population changing in response to earlier evaluation and deployment experience. We advance this as a risk that longitudinal study would be needed to characterise, not as an established mechanism; it is nonetheless the reason to treat verification cost as an indicator monitored over time rather than a figure certified once.}

\viola{\textbf{Latent quantity and operational measures.}} Since deployment risk is governed by whether verification can be completed within an available budget, the budget-relative quantity
\[
\mathrm{Severity}_B^u(x,\hat{y})
=
\frac{\mathbb{E}_{v\sim \rev{\mathcal{V}}}\!\left[C_v^u(x,\hat{y};v)\right]}{B}
\]
\viola{is the quantity of interest, but it is not a quantity a bounded protocol can observe, and neither is its restriction to the budget. An episode that ends at $B$ without a verdict shows only that the cost exceeded the budget, not by how much; an episode that ends early with a mistaken verdict shows neither the cost nor its truncation, since the effort a correct verdict would have required was never expended. What the protocol does observe, in every episode and without exception, is the effort actually spent before the episode terminated. Writing \mbox{$E_d^u$} for that quantity --- the effort expended, in the chosen unit $u$, until termination, whichever of the four outcomes of Step~3 brings it about; for $u=\text{human-minutes}$ it is simply the elapsed verification time --- we report the \emph{observed budget consumption}}
\[
\mathrm{OBC}_B^u(x,\hat{y}) \;=\; \frac{\mathbb{E}_{v\sim\mathcal{V}}\!\left[E_d^u(x,\hat{y};v)\right]}{B} \;\in\; [0,1] ,
\]
\viola{which is a measure of burden rather than of cost, and we name it accordingly: it says how much of the available oversight capacity an output consumes, not how much determining its correctness would require. The primary quantity remains the probability of success, \mbox{$\Pr_{v\sim\mathcal{V}}[S_B(x,\hat{y};v)=1]$}, whose complement enters the definition of the VCE class above. \viola{Neither alone is sufficient: an observed budget consumption near 1 may arise from many episodes ending just short of the budget or from many that exhaust it, \viola{and only the success probability together with its decomposition distinguishes costly but successful verification from termination without a correct verdict.}} Read together they say how much of the available oversight capacity an output consumes, and how often that capacity runs out. The earlier reading of a single ratio crossing unity is thereby retired: \viola{infeasibility is registered by the failure probability within budget, which the protocol observes directly, rather than by a magnitude it cannot.}} Both quantities are defined for any input-output pair \((x,\hat{y})\), but their interpretation depends on output correctness: \viola{for incorrect pairs \viola{the failure probability within budget} is the VCE condition itself, since it is the probability that the error goes undetected within budget, \viola{and the observed budget consumption records how much of the available capacity the attempt absorbed;} \viola{for correct pairs the same two quantities describe the burden of confirmation rather than of error detection.}} \viola{This distinction maps directly onto the stratified reporting introduced in Section~\ref{sec:method}, where outcomes and observed effort are reported separately for outputs that the ground truth marks correct and for those it marks incorrect.}

\rev{\textbf{Complementary failure indicator.}} \viola{An expectation suppresses the heterogeneity of the verifier population, and the part it suppresses is the part that matters for deployment. We therefore report the failure probability
\[
\Pr_{v\sim\mathcal{V}}\!\left[S_B(x,\hat{y};v)=0\right] \;=\; 1-\Pr_{v\sim\mathcal{V}}\!\left[S_B(x,\hat{y};v)=1\right],
\]
the share of the modeled verifier population that does not establish the truth of the matter within budget, and we decompose it into the three ways of failing that Step~3 distinguishes: an incorrect verdict returned within budget, exhaustion of the budget without a verdict, and a declaration that the output is unverifiable. \viola{The three are mutually exclusive and exhaustive of failure, so their rates sum to \mbox{$1-\Pr[S_B=1]$}.} The decomposition is not cosmetic: the first is a failure of the verifier's judgement under a plausible output, the second a failure of capacity, and the third an admission of insufficiency, and they call for different remedies. All three are read off the verification trace directly. The observed budget consumption says how much capacity an output absorbs on average; this indicator says for whom, and in what manner, the capacity proves inadequate.}

\viola{\textbf{On the generation--verification ratio.} The asymmetry motivating this paper invites an obvious index, the ratio of expected verification cost to generation cost, and an earlier formulation of this work carried one. We do not propose it as an indicator, for a reason that is dimensional rather than merely practical. A ratio requires both terms in a common unit, and only a unit that measures machine and human effort alike --- wall-clock time, or money --- satisfies that requirement; under $u=\text{human-minutes}$, which we recommend precisely because deployment reliability is bounded by human oversight capacity, generation has no cost at all and the ratio is undefined. The observation the ratio was meant to capture survives without it, in the qualitative asymmetry of Table~\ref{tab:vbi_asymmetry} and in the regime distinction of Section~\ref{sec:cognitive-problem}. \viola{The quantities we propose for use are therefore the probability of correct verification within budget, the observed budget consumption, and the stratified reporting of Step~5.}}

\viola{Three further features of the verification trace---(i) \textit{time-to-detection}, (ii) \textit{counterexample complexity}, and (iii) \textit{verification length}---are recorded as secondary descriptors of how an episode unfolded. They are not expressed in a common unit and are not interchangeable estimators of a single quantity; they characterise the shape of the verification effort rather than its magnitude.}

{\cfix{A recurring hypothesized pattern in high-verification-cost outputs is that they reproduce surface markers of expertise---structured explanations, domain-specific vocabulary, citations, and authoritative formatting---}\rev{without any guarantee that the content is factually correct or actually supported by the cited evidence}\cfix{; we treat this as a testable hypothesis for verification-aware benchmarking, not as an established result}. This pattern is self-masking at the interface level: each increment in fluency reduces the perceptual cues a verifier can rely on, so verification cost rises with apparent quality. We do not advance a specific training-dynamics mechanism for this pattern; explaining why next-token objectives may generate self-masking outputs remains a topic for future work.}

\viola{Having introduced VCEs and the operational measures that accompany them at a formal level, we now outline a benchmarking methodology that puts them into practice.}

\section{Method: Verification-Aware Benchmarking}
\label{sec:method}

{The proposed method applies to generative systems broadly construed, including Large Language Models (LLMs) and Vision–Language Models (VLMs). Its purpose is not to define a finalized benchmark, but to establish a procedural foundation for benchmarking that explicitly accounts for verification cost. The procedure pairs a standardized reference layer with a recorded \textit{verification trace}, governed by a deployment budget and reported with verifier controls.}

\begin{itemize}
    \item \textbf{Step 1 --- Ground Truth (GT) and Dataset Standardization.} {Evaluation begins with the construction or selection of standardized datasets, together with their associated reference annotations. This reference layer constitutes the Ground Truth (GT), corresponding to the form of ground truth traditionally assumed in benchmark-based evaluation. GT is fixed a priori and shared across all models under evaluation. Its function is to ensure experimental comparability, rather than to guarantee ease of verification.}

    \item \textbf{Step 2 --- Model Execution and Verification Trace (VT).} {Standardized datasets are applied to the generative models under examination. For each model output, verification is performed by a verifier $v$. The aggregated record of these verification activities---verdicts, time-to-verdict, counterexamples produced, and self-reported confidence levels---constitutes the \emph{Verification Trace} (VT) for that output. The VT is not a ground truth: it is a measurement over verifier behavior, distinct from but complementary to GT.} \drift{The record must also fix the generation-side configuration under which the outputs were produced --- model snapshot or version identifier, decoding parameters, and any system-level prompt --- reported alongside the protocol rather than absorbed into it, since without it the resulting indicators cannot be attributed to a determinate system.}

    \item \textbf{Step 3 --- Stopping Rule and Budget.} {Verification proceeds until one of: (a) the verifier reaches a verdict with self-reported confidence $\geq \tau$\footnote{{$\tau$ is a specifically defined threshold, such that $\tau \in \mathbb{R}, 0 \le \tau \le 1$}} (pre-registered, typically $0.9$), (b) the deployment budget $B$ is exhausted, or (c) the verifier declares the output unverifiable within budget. \drift{Outcomes (b) and (c) are recorded as budget-bounded non-verifications. Because a Verification-Cost Error requires the input-output pair to be incorrect (Section~\ref{sec:concept}), such outcomes contribute to the VCE rate only for outputs independently classified as incorrect relative to GT or to the acceptance relation $\mathcal{A}$; for outputs classified as correct \viola{they contribute instead to confirmation-outcome reporting}, and outputs whose status is not independently resolved form a separate indeterminate category. That classification must be independent of the verification episode which the budget interrupted --- GT fixed a priori (Step 1), or an acceptance relation established ex ante --- since otherwise the same bounded process would be used both to detect an error and to establish that there is one.} \viola{This stopping rule is what makes the observed verification effort a finite, well-defined quantity rather than an open-ended audit; it bounds the effort expended, not the effort that establishing correctness would have required.}}
    \viola{Two consequences of the stopping rule must be stated. The first is statistical: when an episode terminates because the budget is exhausted, what is observed is that no correct verdict was reached within $B$; the underlying correct-verdict time is not observed at all. Treating such an episode as though the cost equalled the budget, and averaging it together with completed ones, would bias the resulting figure downward --- systematically, and precisely on the outputs the framework is about. Two terminations must moreover be kept apart, because they are not the same kind of event. Exhausting the budget is administrative censoring: it occurs at a time fixed in advance, independently of the output, and is handled by the usual conventions. A verifier who closes early with a mistaken verdict is not censored at all: the closure is a behavioural outcome that depends on the plausibility of the very output under study, so treating it as censoring would import an independence assumption the framework itself denies. The protocol accordingly records four mutually exclusive outcomes per episode --- correct verdict within budget, incorrect verdict within budget, budget exhausted without verdict, and declared unverifiable --- and reports the probability of the first, \mbox{$\Pr[S_B=1]$}, as the primary quantity. \viola{Reporting that probability as a function of $B$, rather than a single threshold crossing, also avoids the information loss inherent in any binary cut. Such a curve, however, is not obtained by re-reading a trace collected under one budget: the announced budget may itself shape how a verifier allocates attention, so varying $B$ does not merely truncate a fixed process. Comparisons across budgets require separate or randomised budget conditions, and a trace gathered under one declared budget supports the primary estimate for that budget alone.} The second consequence is behavioural: the rule is exposed to a bias this paper itself documents. If fluent, authority-mimicking outputs induce unwarranted confidence in those who read them (Sections~\ref{sec:broken}--\ref{sec:limits}), a verifier may reach $\tau$ early on precisely the outputs whose defects are most expensive to expose, so that decision time, if mistaken for verification cost, would make the burden appear artificially low where it matters most. Inter-rater agreement does not detect this: a heuristic shared across the verifier population produces confident \emph{agreement}, not disagreement, and would raise rather than lower the agreement statistic of Step 4. The protocol therefore requires that the verdicts recorded in the VT be compared against GT, and that the rate of \emph{premature closures} --- verdicts reaching $\tau$ yet disagreeing with GT --- be \viola{reported alongside the operational measures}. This rate is not merely a control on the instrument: an output that induces a confident and incorrect closure is the limiting case of the failure class this paper is concerned with, and measuring it is arguably the most informative thing the protocol does.}

    \item \textbf{Step 4 --- Verifier Control.} {\viola{To separate system-associated verification burden from verifier-skill variance}, every protocol instantiation must include (i) a verification baseline measured on a held-out set of known-correct outputs, (ii) inter-rater agreement statistics (e.g., Krippendorff's $\alpha$) across $N \geq 3$ verifiers per output, and (iii) verifier-experience metadata that supports stratified analysis.
    \viola{A population-level reading of the quantities defined in Section~\ref{sec:concept} additionally requires that the verifiers actually observed represent the declared distribution $\mathcal{V}$. Recruitment, sampling or stratification design, eligibility criteria, and the extent of repeated participation across outputs are therefore part of what is pre-registered. Where verifiers are not sampled representatively, the estimates are to be read as statements about the observed verifier panel and not about a wider population; and where the same verifiers assess many outputs, the model-level uncertainty must account for dependence within verifier, since treating such observations as independent understates it.} \viola{Model-level indicators are reported in all cases, and accompanied by the agreement statistic rather than gated on it: results obtained below a pre-registered agreement threshold are flagged as ``verifier-uncertain'' and interpreted accordingly, but they are not withheld, since low agreement may itself be evidence that verification is difficult.}}
    \viola{Agreement alone, however, is a weak guarantee, and using it as a gate is weaker still. As Step~3 notes, a heuristic shared across the verifier population produces confident agreement on exactly the outputs the framework targets, so a high coefficient may certify a systematically mistaken reading; conversely, a legitimately ambiguous task may be discarded for low agreement while carrying real information about verification difficulty. Three quantities should therefore be reported side by side rather than collapsed into an admission criterion: inter-rater agreement, verdict accuracy against GT, and the rate of premature closures. The agreement threshold governs how confidently a model-level aggregate may be stated; it does not establish that the aggregate is correct.}
    \rev{This separation is partial, not complete: the known-correct baseline estimates a task- and verifier-dependent \viola{observed \emph{confirmation} effort}, and the excess observed on model outputs --- under a fixed dataset, interface, tool access, verifier population, and protocol --- is reported as a comparative, model-associated increment rather than a causal attribution, since confirming correct outputs and refuting incorrect ones may involve different procedures (Step 5).}

    \item \textbf{Step 5 --- Stratified Reporting.} \viola{Verification outcomes and observed effort are reported separately for GT-correct and GT-incorrect outputs, since confirming a correct output and refuting an incorrect one are distinct cognitive tasks with distinct effort profiles.} \viola{The operationally relevant pair for VCE detection is the probability of correct refutation within budget, \mbox{$\Pr[S_B=1 \mid \text{GT-incorrect}]$}, together with the observed budget consumption on that stratum; the corresponding expectations over \mbox{$C_v^{u}$} are not reported, since the protocol does not observe them. \viola{Time to a correct verdict is reported only over the episodes in which one was reached, and always labelled \emph{conditional correct-verdict time among successful episodes}, so that it is not mistaken for an estimate of the full distribution of the latent cost.}}
    \bridge{Because the two strata are computed over disjoint sets of outputs, they yield indicators of different status.} \viola{The observed budget consumption varies with $B$ and therefore measures oversight effort against the capacity available in a deployment; the ratio of consumption across the two strata, \mbox{$\mathbb{E}[E_d^u \mid \text{GT-incorrect}]/\mathbb{E}[E_d^u \mid \text{GT-correct}]$}, instead characterises how much more effort refutation absorbs than baseline confirmation. The ratio is dimensionless, the unit $u$ having cancelled, but it is not thereby budget-independent: $B$ does not appear in it algebraically, yet it shapes the \viola{observed effort}, since a larger budget allows episodes to run on where a smaller one would have terminated them. It is therefore comparable only under a common budget and otherwise harmonised protocols. \viola{Both quantities are read off observed effort and require no counterfactual.}} \bridge{ Governance frameworks that carry human oversight as an explicit cost term rather than as an implementation detail \citep{Immediato2026-IJSC} require precisely this distinction between the \emph{volume} of oversight and its \emph{difficulty}; reporting a single aggregate figure discards it.\footnote{\bridge{Concretely, \citet{Immediato2025-POM} decomposes that cost term as $\eta_1 \cdot \text{CognitiveVerification} + \eta_2 \cdot \text{EthicalOversight}$, where $\eta_1$ and $\eta_2$ scale the respective burdens, and \citet{Immediato2026-IJSC} restates the aggregate as an oversight workload $H$ under a complexity factor $\eta$. Since the present paper adopts \emph{Cognitive Verification} in that sense (Section~\ref{sec:cognitive-problem}), the budget-relative quantity is an estimator of the cognitive-verification burden alone; the ethical-oversight term lies outside the scope of this protocol. Whether the \viola{dimensionless but budget-dependent observed-effort} ratio is in turn a usable estimator of the corresponding complexity factor $\eta_1$ is a further question, recorded here as a proposal for downstream work rather than as a result of this paper.}}}

    \item \textbf{Step 6 --- Model-Level Reporting.} {For each generative model, \viola{verification outcomes and observed-effort estimates derived from VT are aggregated} to yield \viola{the model-level observed budget consumption \mbox{$\mathbb{E}[E_d^u]/B$}, the probability of a correct verdict within budget, \viola{and the confirmed \mbox{$\mathrm{VCE}_{B,q}$} rate together with the inconclusive rate and the resulting classification envelope,} \viola{reported together with the stratified outcomes and observed effort of Step~5.} These are not intended as universal scalar metrics; they are comparative indicators that reflect how strongly a given system shifts verification effort onto external evaluators.}}
    \viola{Where comparison must span deployments differing in cost unit, the ratio \mbox{$\mathbb{E}[E_d^u \mid \text{GT-incorrect}]/\mathbb{E}[E_d^u \mid \text{GT-correct}]$} of Step~5 is dimensionless and invariant under a common multiplicative rescaling of the chosen effort unit --- minutes to seconds, say --- but it is not invariant to a change in the underlying cost representation, since moving from human-minutes to a monetary unit reweights verifiers whose time is priced differently. It is not, however, a domain-free quantity, and we do not offer it as one: \viola{the budget shapes the observed effort it is built from}, and task difficulty, error prevalence, the difficulty of establishing ground truth, tooling and verifier expertise continue to shape both strata. Comparison across deployments requires a common budget and otherwise harmonised protocols; absent those, no quantity proposed here transports.}
\end{itemize}

\rev{\textbf{Illustrative instantiation (non-estimative).} As a purely illustrative walk-through, consider code generation. GT is a curated set of tasks with reference test suites (Step 1); the specification fixes $u=\text{human-minutes}$, $\mathcal{V}$ as a population of professional developers with declared seniority metadata, $B=15$ human-minutes per output, $\tau=0.9$, \viola{$q=\tfrac12$, Clopper--Pearson intervals at the $95\%$ level, and the three-state decision rule of Section~\ref{sec:concept}, \viola{with the number of verifiers per output, $m$, fixed by a precision-planning step so that the tolerated share of inconclusive classifications is met, rather than set at the Step~4 minimum}}. For each model output, a verifier attempts confirmation or refutation until the stopping rule triggers, and the VT records the verdict, time-to-verdict, and any counterexamples (Steps 2--3); \viola{observed budget consumption, success probability, and the confirmed and inconclusive VCE rates with their envelope are then aggregated under the controls of Steps~4--6.} All values above are specification choices, not findings: the walk-through shows how the constructs are instantiated and carries no empirical weight.}

{\textbf{Domain considerations.} The deployment context shapes not only the values of GT, $\mathcal{V}$, and $B$, but also \viola{the verification-outcome profile of VCEs}. In safety-critical or high-accountability domains, $B$ is structured by regulatory and accountability regimes, and a VCE can propagate into physical, legal, or institutional harm. In such settings, the relevance of the protocol is not merely methodological: adherence to structured verification practices carries direct industrial significance, as failures to detect VCEs may escalate from localized model errors into system-level risk, regulatory exposure, and operational harm.
In organizational domains, $B$ is shaped by internal processes, and undetected VCEs may accumulate as technical or epistemic debt. In open-ended creative domains, by contrast, $B$ is often more elastic, and the cost of an undetected plausible error is more likely to be reputational, interpretive, or editorial than immediately physical or safety-critical. We do not claim that the framework applies uniformly across these regimes; rather, the protocol's parameters \viola{$(\mathcal{V}, B, \tau, q)$} are the loci at which domain-specific risk profiles enter the evaluation.}

\viola{\textbf{Cost of instrumentation.} The protocol is not free, and its cost is dominated by verifier time rather than compute. \viola{If $m$ verifiers assess each of $n$ outputs and Step~3 bounds each episode by $B$, the protocol consumes at most $mnB$ \viola{units of verifier effort --- verifier-minutes when $u=\text{human-minutes}$ ---}, plus the baseline and calibration overhead of Step~4. The requirement $m\ge 3$ supports agreement analysis; output-level classification may require a larger $m$, as Section~\ref{sec:concept} notes, and the instrumentation cost scales linearly with it.} Two consequences follow. First, verification-aware evaluation is affordable at the scale of a curated, stratified subset rather than an exhaustive benchmark, and should be designed as such. Second, because the budget bounds each episode by construction, the total cost is bounded in advance: this is what distinguishes the procedure from an open-ended audit and makes it schedulable within an evaluation campaign.}

{The method yields a verification-aware benchmarking procedure in which models can be compared not only by accuracy with respect to GT, but by the verification burden revealed through VT. \viola{The procedure is a domain-adaptable human-verification protocol: GT, $\mathcal{V}$, and $B$ are chosen to fit the deployment context}, while the structural requirements---recording verification effort against a deployment budget, with verifier controls and stratified reporting---are fixed across instantiations. We leave concrete instantiations and empirical evaluation to future work.}
\rev{The protocol is thereby designed to support statistical replication relative to a declared, pre-registered measurement specification $(u,\mathcal{V},B,\tau,\viola{q})$: variability across verifiers is modeled and reported rather than eliminated, following the logic of replication used in many human-subject measurement protocols.}
\drift{Because that specification may itself drift, every reported result should additionally carry the date of measurement and the versions of the protocol and of the evaluated system configuration under which it was obtained, and instantiations should be repeated across releases rather than performed once. \viola{Verification outcomes and burden are accordingly proposed as indicators to be tracked over time}, in the way reliability indicators are tracked in operational engineering practice, rather than as a certificate issued at a single point.}

\section{Implications: Evaluation and System Design}
\label{sec:implications}

Introducing verification cost as a primary metric changes both evaluation and system design. Optimizing for correctness alone is insufficient when verification is expensive.

\paragraph{Evaluation.} \tidy{Measured under constrained budgets, verification-aware evaluation separates errors that are trivial to detect from those that require substantial effort: in code generation it extends beyond test pass rates to the effort required to expose failures, and in factual generation to the effort required to validate claims against external sources.}

Evaluation should also target outputs that are difficult to falsify. Adversarial benchmarking exposes model weaknesses \citep{kiela2021dynabench}, but does not prioritize high-verification-cost outputs. A verification-aware benchmark focuses on these cases, aligning evaluation with real deployment risk.

\paragraph{System Design.} Systems should minimize verification burden, not only error rates. This requires designing outputs that support efficient validation. One approach is to generate intermediate representations that are independently checkable, such as executable programs or structured claims.
However, additional structure must reduce validation effort rather than shift it.
Explanations should be evaluated by how they affect verification cost: they are valuable only when they reduce the effort needed to validate outputs.
Uncertainty estimation should guide verification rather than signal correctness.

\section{Alternative Views}
\label{sec:alternative}

\rev{A common assumption is that reliability can be improved within existing paradigms, treating correctness as primary and verification as derivative. We develop the three strongest alternatives to our position and state, for each, why it leaves the cost dimension unmeasured despite its merits.}

\rev{\textbf{Graded correctness and reward modeling.} The most significant alternative argues that the correctness score itself should change: rather than supplementing a binary verdict with a cost metric, the evaluator should assign a continuous score that reflects how effectively core functionalities and edge cases are handled---a view consistent with reward modeling in RL optimization pipelines. We regard this alternative as complementary rather than competing. Moving correctness onto a continuous scale changes the codomain of the grader's verdict; our dimension is the cost of \emph{producing} that verdict, regardless of its scale. A graded score must, however, be determined by a verifier within a budget, and grading edge-case coverage is often the most time-consuming part. The two proposals complement each other naturally: a continuous grading scale presented alongside a verification trace.}

\rev{\textbf{Calibration and selective prediction.} A second alternative holds that models should flag their own unreliable outputs, concentrating human effort where confidence is low \citep{jiang2021calibration,yin2023selfaware}. Calibration indeed helps allocate verification effort, but it relies on an epistemic self-assessment that models often lack, and it does not lower the per-output cost of verification: high-stakes outputs demand audit regardless of reported confidence. Selective prediction reduces how often verification happens, not what it costs when it happens.}

\rev{\textbf{Benchmark-centric improvement.} A third alternative holds that richer datasets and holistic or adversarial evaluation will progressively capture the remaining failure modes \citep{liang2023helm,kiela2021dynabench,kapoor2026hal}. These efforts broaden \emph{what} is measured, but continue to measure agreement with references while leaving unrecorded the labor required to establish that agreement; in deployment, correctness often requires external reasoning, domain expertise, or cross-source validation. Tool-based pipelines \citep{gao2023pal} similarly improve correctness while adding components whose outputs must themselves be validated, making verification multi-stage rather than cheaper.}

\rev{Each alternative improves what is measured; none measures the cost of measuring. Reliability fails not only when models are wrong, but also when errors cannot be detected efficiently, and evaluation remains misaligned as long as it accounts for correctness without accounting for verification cost.}

\section{Concluding Remarks}
\label{sec:concludingremarks}
\rev{\viola{This article argues that correctness alone is not a sufficient metric for measuring reliability and that the \viola{AI evaluation} should report not only correctness but also the verification results and the observed effort, \viola{in relation to a stated budget for verifying the implementation}. \viola{The verification cost itself remains the latent target}; what a bounded protocol can deliver are indicators of it.} To this end, we introduced Verification-Cost Errors as \viola{the class of incorrect input-output pairs that \viola{a specified fraction} of the verifier population fails to identify within that budget}; \viola{\viola{an observed budget-consumption measure}, supplemented by the probability of correct verification within budget and by its decomposition into the various ways in which verification fails, as operational metrics;} and we have also introduced a benchmarking protocol based on a six-step verification process, featuring auditor checks and tiered reporting. These instruments are deliberately conceptual: verifier variance, unit choice, and the acceptance criterion are declared parameters of the measurement --- not difficulties that we claim to have eliminated. \drift{Four} directions follow. First, a controlled instantiation of the protocol that compares \viola{systems} with similar accuracy in terms of measured verification burden. Second, standardized reporting norms --- \viola{\viola{success probability within budget, observed budget consumption, and the confirmed and inconclusive VCE rates alongside accuracy}} --- \viola{so that verification outcomes and observed burden become routine columns of evaluation tables}. Third, integration with related programs: graded correctness and reward modeling, the emerging science of agent reliability \citep{rabanser2026reliability}, \bridge{governance metrics that carry an explicit human-oversight cost term, for which the stratified quantities of Step 5 would supply measured rather than stipulated inputs \citep{Immediato2026-IJSC},} and the design of outputs that are cheap to verify by construction. \drift{Fourth, longitudinal measurement: since verifier populations, tooling, budgets, and the deployed systems themselves may all drift, what matters most is \viola{the trajectory of verification burden across successive releases}, rather than a single measurement. Whether this trend is increasing, decreasing, or remaining stable as the systems become more accurate is an empirical question that can only be answered through \viola{verification-aware} benchmarking conducted over the long term.} If the purpose of the evaluation is to predict the deployment risk, it is necessary to measure the cost of verifying whether a result is correct.}

\bibliographystyle{plainnat}
\bibliography{references}
\end{document}